%% file: topic_segmentation_ecir.tex
\documentclass[runningheads,a4paper]{llncs}

\usepackage{amssymb}
\usepackage{graphicx}

\usepackage{booktabs} 
\usepackage{amsmath}
\usepackage{multirow}
\usepackage{graphicx}
\usepackage{tikz,times} 
\usepackage{pgfplots}
\usepackage{wrapfig}	
\usepackage{subcaption}
\renewcommand\textfloatsep{4pt}

\renewcommand\intextsep{4pt}

\DeclareMathOperator*{\argmax}{arg\,max}

\DeclareMathOperator{\softmax}{softmax}

\begin{document}

\mainmatter  

\title{Attention-based Neural Text Segmentation\vspace{-20pt}}
\author{}\institute{}
\author{Pinkesh Badjatiya \and Litton J Kurisinkel\and Manish Gupta\thanks{The author is also a Principal Applied Scientist at Microsoft}\and Vasudeva Varma}
\institute{IIIT-H, Hyderabad, India\\
\{pinkesh.badjatiya, litton.jkurisinkel\}@research.iiit.ac.in,  \{manish.gupta,vv\}@iiit.ac.in}
\authorrunning{Badjatiya et al.}

%
%

\maketitle

\input{abstract}
\input{introduction}
\input{relatedwork}
\input{proposed_method}

\input{experiments}
\input{results}

\input{conclusion}



\end{document}

%% file: abstract.tex
\begin{abstract}
Text segmentation plays an important role in various Natural Language Processing (NLP) tasks like summarization, context understanding, document indexing and document noise removal. Previous methods for this task require manual feature engineering, huge memory requirements and large execution times. To the best of our knowledge, this paper is the first one to present a novel supervised neural approach for text segmentation. Specifically, we propose an attention-based bidirectional LSTM model where sentence embeddings are learned using CNNs and the segments are predicted based on contextual information. This model can automatically handle variable sized context information. Compared to the existing competitive baselines, the proposed model shows a performance improvement of $\sim$7\% in WinDiff score on three benchmark datasets.
\end{abstract}

%% file: introduction.tex
\section{Introduction}
\label{sec:introduction}
The task of text segmentation is defined as the process of segmenting a chunk of text into meaningful sections based on their topical continuity. 
Text segmentation is one of the fundamental NLP problems which finds its use in many tasks like summarization~\cite{mitrat1997automatic}, passage extraction~\cite{oh2007semantic}, discourse analysis~\cite{van1982episodes}, Question-Answering~\cite{oh2007semantic}, context understanding, document noise removal, etc. Fine grained segmentation of a document into multiple sections provides a better understanding about the document structure which can also be used to generate better document representations, which in turn could benefit other natural language applications. 
Complexity of text segmentation varies with the type of text and writing styles -- informational, conversational, narrative, descriptive, etc. In some cases, context is a very important signal for the task, while in other cases, dependence on context may be minimal. Also, complex topic shifts in the text and use of abstract cue phrases in the sentences make the task challenging.

%

Multiple supervised and unsupervised methods have been already proposed to tackle some of these challenges. Many unsupervised methods are heuristic and ad hoc in nature, need huge memory, have large execution times, and do not generalize well across multiple text types. Supervised methods require labeled data and often the performance of such systems comes at the cost of hand-crafted highly tuned feature engineering. None of these methods can automatically tune the degree of dependence on the context. 
Sequence-to-sequence models like Recurrent Neural Networks (RNNs) and Long Term-Short Memory (LSTMs) can model sequences effectively by controlling information flow across time. Such models can in general help capture long range dependencies (context) but they work well with short sequences. They can be enhanced by giving varying attention weights to sentences in the context, where the weight denotes the relative importance of a context sentence for segmentation. Attention thus allows us to learn the focus points from the context. To the best of our knowledge, this is the first work to explore the use of attention-based neural mechanism for text segmentation. 
We propose a novel Attention-based CNN-BiLSTM model that learns to represent the context of the sentence by learning the attention weights. The proposed model does not require any manually designed features, is domain independent and scalable. 
The proposed neural model architecture is illustrated in Figure~\ref{fig:model-architecture}. We compare the proposed method with competitive baselines on three benchmark datasets.

In Section \ref{sec:related-work}, we review the existing work on text segmentation. Section \ref{sec:proposed-method} describes the proposed Neural model with Attention-based approach. Section \ref{sec:experiments} compares the performance of various methods on benchmark datasets. In Section \ref{sec:results}, we analyze the results and  conclude with a brief summary in Section \ref{sec:conclusion}.

%% file: relatedwork.tex
\vspace{-5pt}
\section{Related Work}
\label{sec:related-work}

Unsupervised methods for text segmentation include lexical cohesion~\cite{eisenstein2008bayesian,hearst1997texttiling}, statistical modeling~\cite{beeferman1999statistical,utiyama2001statistical}, affinity propagation based clustering~\cite{kazantseva2011linear,sakahara2014domain}, and topic modeling~\cite{eisenstein2008bayesian,misra2009text,Purver:2006:UTM:1220175.1220178,riedl2012topictiling}. Topic modeling approaches include PLDA~\cite{Purver:2006:UTM:1220175.1220178} (captures the amount of topic distribution that a paragraph shares with its predecessor), SITS~\cite{nguyen2012sits} (chains a set of Hierarchical Dirichlet Process LDAs), TSM~\cite{du2013topic} (integrates point-wise boundary sampling with topic modeling), and~\cite{du2015topic} (ordering-based probabilistic topic models to incorporate the ordering irregularity into the probabilistic approach). These methods are globally informed, i.e., they consider the whole document when generating the most probable segment boundaries. However, huge memory requirements and large execution times make these methods unpractical for use in real applications.

Various classifiers like decision trees~\cite{grosz1992some,tur2001integrating} and probabilistic models~\cite{beeferman1999statistical,hajime1998text,reynar1999statistical,utiyama2001statistical} have been proposed for supervised text segmentation. Popular features include lexical (like lexical similarities~\cite{hearst1997texttiling}), conversational (acoustic indicators, long pauses, shifts in speaking rates, higher maximum accent peak, cue phrases, silences, overlaps, speaker change~\cite{galley2003discourse}) and knowledge-based features~\cite{joty2011supervised}. Supervised methods require labeled data and hand-crafted highly tuned feature engineering. Also, they are locally informed and often fail to capture the overall global topic structure of the document. 

Some previous studies, although scarce and somewhat preliminary, have explored neural approaches for domain-specific text segmentation. Sheikh et al.~\cite{sheikh2017topic} proposed a method for segmentation in transcripts using RNNs, Wang et al.~\cite{wang2017learning} attempt to learn a coherence function using the partial ordering relations, Wang et al.~\cite{wang2016topic} use BiLSTM-CNN to model the task as a simple binary classification task for Chinese. In this paper we explore the use of attention-based deep neural architecture for the task of automatic linear text segmentation, which provides a good trade off between the locally informed and globally informed behavior by varying the amount of context information used.

%% file: proposed_method.tex
\section{The Proposed Method}
\label{sec:proposed-method}
In this section, we start by presenting the formal problem definition. Further we discuss steps related to the data preparation and pre-processing. Finally, we present our neural model architecture.

\subsection{Problem Definition}

We model the text segmentation problem as a binary classification problem. Given a document, we define the problem with respect to the $i^{th}$ sentence in the document, as follows.

\noindent\underline{Given}: A sentence $s_i$ with its $K$ sized \textit{left-context} $\{s_{i-K},\ldots,s_{i-1}\}$ (i.e., $K$ sentences before $s_i$) and $K$ sized \textit{right-context} $\{s_{i+1},\ldots,s_{i+K}\}$ (i.e., $K$ sentences after $s_i$). Here $K$ is the context size.

\noindent\underline{Predict}: Whether the sentence $s_i$ denotes the beginning of a new text segment. 

In this paper, we propose a neural framework to tackle this problem. Using a neural framework, we aim at using the context for learning distinctive features for sentences that mark the beginning of the segment. The architecture of the proposed model is illustrated in Fig.~\ref{fig:model-architecture}.

\subsection{Data Preparation}
In this section, we discuss two main steps in data preparation: pre-processing and custom batch creation to incorporate neighboring context.

\subsubsection{Data Pre-processing}
\label{proposed_method:pre_processing}

We fix the length of sentences to $L$ words and truncate/pad as required to achieve appropriate fixed length embedding of the sentences. To represent words, we use the 300D word2vec\footnote{https://code.google.com/archive/p/word2vec/} embeddings which are trained on Google News dataset containing $\sim$100B words with a vocabulary size of $\sim$3M words.

Let $V$ represent the vocabulary, and let $d$ be the word embedding size. Let $E^{V \times d}$ be the embedding matrix whose each row represents the embedding of a particular word in the model vocabulary. Let $\eta_i$ be a matrix whose $j^{th}$ row corresponds to the one-hot representation of the $j^{th}$ word of the sentence $s_i$. Thus, $\eta_i$ has $L$ rows and $V$ columns. Given the word embedding matrix $E$, we obtain the sentence embedding matrix $e(s_i)$ for sentence $s_i$ as $e(s_i)^{L \times d} = \eta_{i}^{L \times V} \times E^{V \times d}$. 

While creating $\eta_i$ for a sentence $s_i$, we perform basic text cleaning steps like skipping the punctuations and stop words. For all the missing words in the word2vec vocabulary we perform WordNet-based lemmatization and use the lemmatized word instead. If the embedding is still missing from the vocabulary then we replace it with a special token $\langle$UNK$\rangle$.

\subsubsection{Custom Batch Creation}
\label{proposed_method:neighboring_context}

We wish to exploit the context around a sentence to decide whether the sentence indicates a segment boundary. For a sentence $s_i$, let \textbf{$lc_{i}$} and \textit{$rc_{i}$} be the one-hot representations of the \textit{left-context} and \textit{right-context} respectively. Both $lc_i$ and $rc_i$ therefore contain $K \times L$ words each. Their embeddings \textbf{$e(lc_{i})$} and \textbf{$e(rc_{i})$} can then be computed as $e(lc_{i})^{K \times L \times d} = lc_{i}^{K \times L \times V} \times E^{V \times d}$ and $e(rc_{i})^{K \times L \times d} = rc_{i}^{K \times L \times V} \times E^{V \times d}$ respectively. Note that we also refer to the sentence $s_i$ as the \textit{mid-sentence}. We perform padding as required to obtain a fixed length representation of size $K\times L\times d$ for both the contexts. The input to the model is a batch of samples with the $i^{th}$ sample, $S_i$, defined as the concatenation of the embeddings of the \textit{left-context}, \textit{mid-sentence} and the \textit{right-context} as follows.
\vspace{-5pt}
\begin{eqnarray}
	S_i = [e(lc_i)^{K \times L \times d}, e(s_i)^{L \times d}, e(rc_i)^{K \times L \times d}]
\end{eqnarray}
\vspace{-5pt}

\begin{figure}%
\centering
\includegraphics[width=0.8\columnwidth]{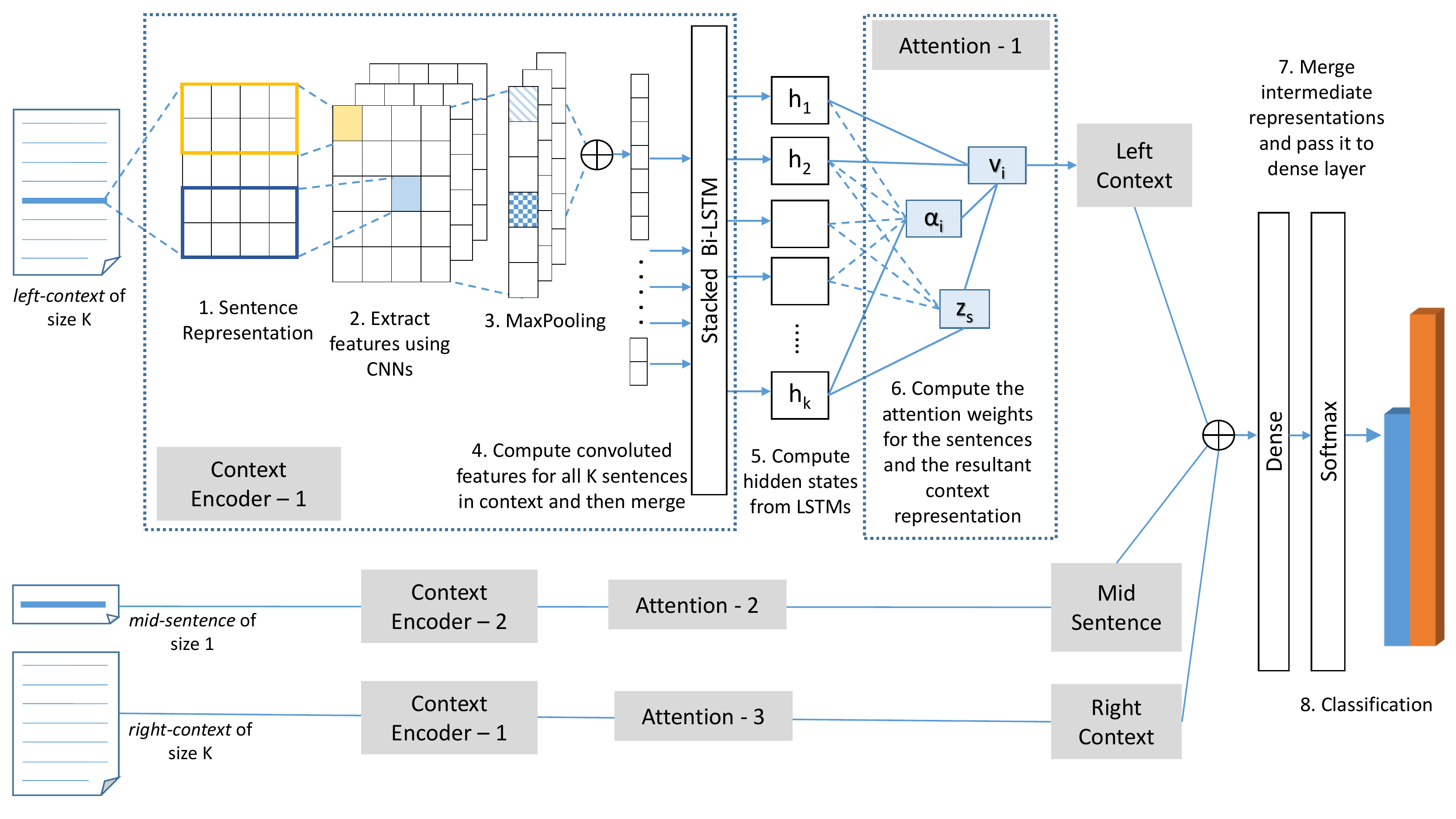}%
\caption{Architecture Diagram for the Proposed Model}%
\label{fig:model-architecture}%
\end{figure}

The context size $K$ should be such that the covered neighborhood information is enough to make conclusive decision about the current sentence being a segment boundary or not. Higher values of $K$ provide the model with extra unnecessary context along with increase in the number of parameters. Lower values of $K$ reduces the model complexity, but also restricts the model's ability to capture relations across near sentences only. $K$ can be tuned using validation data. We study sensitivity of results to variation in $K$ in Section~\ref{sec:results}.

\subsection{The Proposed Neural Model}
\label{proposed_method:neural_model}
We now discuss in detail the proposed neural network model as illustrated in Fig.~\ref{fig:model-architecture}. The data pre-processing and sentence embedding discussed in the previous sub-section provides us instances of the form $S_i$, which consist of \textit{left-context}, \textit{right-context} and the \textit{mid-sentence} word-embedding representations. This corresponds to the output at Step 1 in Fig.~\ref{fig:model-architecture}.

\subsubsection{CNN Transformations}

We leverage the widely used CNN architecture to obtain rich feature representations for each sentence in the \textit{left-context}, \textit{mid-sentence} as well as the \textit{right-context}. Recall that the embedding $S_i$ has $2K+1$ rows each having $L\times d$ dimensions. Let us denote the $j^{th}$ such embedding matrix in $S_i$ as $S_{(i,j)}^{L\times d}$. On each such $S_{(i,j)}$, we perform 1D Convolution operations with $z$ number of filters. Let us denote the $l^{th}$ filter with a set of weights $\{\omega_l^{h \times d}, b_l\}$. Such a filter with height \textit{h} can be applied on the input $S_{(i,j)}$ to obtain feature maps as follows. 
\vspace{-10pt}
\begin{eqnarray}
f_{kl} =  \phi(\omega_l^{h \times d} \cdot S_{(i,j)}[k-\frac{h}{2}: k+\frac{h}{2}]^{h \times d} + b_l)
\end{eqnarray}
\vspace{-15pt}

Note that the convolution operations on text data involve filters with width same as input dimensionality ($d$). Thus, a filter has dimensions $d\times h$. Here, $f_{kl}$ denotes the result of the convolution using a non-linear transformation $\phi$. The filter is applied to each row of $S_{i,j}$. After applying $z$ such filters, for each row $k$ of $S_{i,j}$, we obtain a feature vector $f_{k} = \{f_{k1}, f_{k2}, \dots, f_{kz}\}$. This corresponds to the output at Step 2 of Fig.~\ref{fig:model-architecture}. 

Max-pooling is a popular sample-based discretization operation in CNNs. Given the feature vector $f_{k}$, max pooling operation involves computing the maximum feature value per filter across a group of rows in $S_{i,j}$. We pool across all the $L$ rows in $S_{(i,j)}$ and get one value per filter (or feature map). We perform this operation for all the filters. Thus, overall, we obtain a feature rich representation of size $z$ per sentence in $S_{(i,j)}$. We perform the convolution operation for all the sentences independently and obtain context representation by concatenating the sentence representations in the same sequence. Recall that $S_i$ contained representations of $2K+1$ sentences. Thus, overall the instance $S_i$ is now represented by a sequence $TS_i$ with $2K+1$ units each of size $z$. $TS_i$ is the output at Step 4 of Fig.~\ref{fig:model-architecture}.

We use shared filters for the \textit{left-context} and \textit{right-context} because: (1) It reduces the number of trainable parameters drastically making it easier for the model to train. (2) The representation vectors generated for both the \textit{right-context} and \textit{left-context} have the same semantics and lie in the similar vector space.

\subsubsection{Stacked BiLSTMs with Attention}

The problem could have been modeled as a sequence to sequence label generation task where each training sample is a whole document. But this model would be difficult to generalize for variable document length. Also, LSTMs have been shown to work well for shorter sequences. Hence, we first used CNNs to generate sentence embeddings and then use BiLSTM network on a smaller sequence that consists of only the main sentence and its neighbors. 

To obtain a unified rich feature representation, we use Attention Bidirectional Long-Short Term Memory Network (Attention-BiLSTM) on top of this sequence $TS_i$. LSTMs~\cite{hochreiter1997long} have been shown to model sequences better than vanilla Recurrent Neural Networks (RNNs) for various NLP tasks. LSTMs keep memories to capture long range dependencies. These memory cells allow error messages to flow at different strengths depending on the inputs. LSTMs have the ability to control the flow of information that flows to the memory cell state by using structures called \textit{gates}. The reader is referred to~\cite{hochreiter1997long} for details about LSTMs. To obtain a unified context representation which has rich feature set, we use a bidirectional LSTM (BiLSTM). The resultant embeddings are the concatenation of the two embeddings obtained through a forward pass LSTM and a reverse pass LSTM, capturing information from both the directions. 

As shown in Fig.~\ref{fig:model-architecture}, we model the sequence of size $K$ for both the left and the right context parts of $S_i$ using separate LSTM networks each having $K$ such memory cells, intuition being, two sentences at equal distance from the middle sentence might not have similar effect on the sentence being a segment boundary.

Traditional sequence encoder architecture which uses stacked BiLSTMs, forces the encoder to capture the information in a single fixed length representation. This also has a drawback as the hidden state at $h_t$ is dependent on $h_{t-1}$ across consecutive layers in the stack, thus the hidden state $h_0$ will have a significant effect on the future states. To overcome this, we use two vertically stacked BiLSTMs followed by soft attention~\cite{xu2015show}. 
Attention allows the model to give more importance to certain set of sentences in the context while ignoring the others, effectively learning the focus points to better predict if a sentence forms a segment boundary. We introduce an attention vector, $z_s$ and use it to measure the relative importance of the sentences in the context as follows. Let $H_i^{K \times sz}$ denote the output of the last BiLSTM layer. Here $sz$ is the size of the BiLSTM output corresponding to a sentence in the context. As shown in Fig.~\ref{fig:model-architecture}, $H_i=\{h_1, \ldots, h_K\}$.

\vspace{-15pt}
\begin{eqnarray}
	& e_{i}^{K \times 1} = H_i^{K \times sz} \times W^{sz \times 1} + b_{i}^{K \times 1} \\
	& a_i = \exp(\tanh(e_{i}^T z_s)), \ \ \  \alpha_{i} = \frac{a_{i}}{\sum_{p} a_{p} }, \ \ \ v_i = \sum_{j=1}^{K} \alpha_j h_j
\end{eqnarray}
\vspace{-10pt}

The resultant context embedding, $v_i$, which is jointly learned during the training process captures the essential information from the context. We compute this context embedding for both the \textit{left-context} and \textit{right-context}. The merged vector corresponds to the output at Step 7 of Fig. \ref{fig:model-architecture}.

\subsubsection{Dense Fully Connected Layer with Softmax}

The resultant embeddings obtained from a shared encoder for the \textit{left-context} and \textit{right-context} and the embedding for \textit{mid-sentence} obtained from separate but similar encoder are passed to a dense fully connected layer followed by an output.

\vspace{-15pt}
\begin{eqnarray}
P(y|s_i) = \softmax(W_h \times v_i + b_i), \ \ \ \  y_{pred} = \argmax\ P(y|s_i)
\end{eqnarray}
\vspace{-10pt}

For classification we have a softmax layer over the output vectors. Finally, we take $\argmax$ over the predicted probability distribution to generate predictions.

%% file: experiments.tex
\section{Experiments}
\label{sec:experiments}

In this section, we discuss datasets, metrics and parameter settings for our experiments. Source code and datasets are available at \url{https://github.com/pinkeshbadjatiya/neuralTextSegmentation}.

\subsection{Datasets}

\begin{figure}[!htb]
    \centering
    \begin{minipage}{.55\textwidth}
	\centering
	\scriptsize
		\begin{tabular}{|p{0.65in}|l|l|c|c|}
			\hline
			\multicolumn{2}{|l|}{}& \bf Clinical &  \bf Fiction  &  \bf Wikipedia \\ \hline
			\multicolumn{2}{|l|}{\#Documents} & 227 & 85  & 300 \\ \hline
			\multicolumn{2}{|l|}{\#Sentences or \#Samples} & 31868 & 27551  & 58071 \\ \hline
			\multirow{2}{*}{Segment Length} & Mean & 35.72 & 24.15 & 25.97 \\ \cline{2-5}
			& Std Dev & 29.37 & 18.24 & 9.98 \\
			\hline
		\end{tabular}
	\caption{\label{data-stats-table} \scriptsize Statistics of Various Datasets used for Performance Evaluation}
    \end{minipage}%
    \begin{minipage}{0.4\textwidth}
		\scriptsize
\includegraphics[width=\columnwidth]{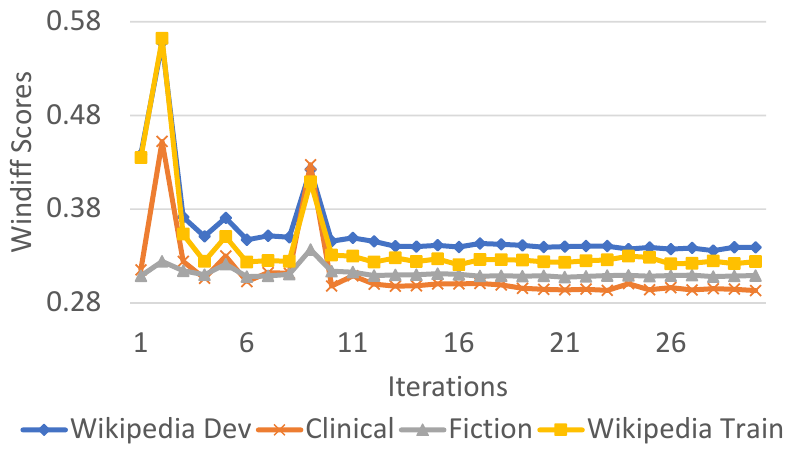}%
\caption{\scriptsize Variation of WinDiff scores wrt Iterations for Various Datasets}%
\label{fig:windiffIters}%
    \end{minipage}
\end{figure}

Text segmentation is quite a subjective task making evaluation of the text segmentation systems challenging. Hence, we use standard benchmark datasets for evaluation. Table \ref{data-stats-table} shows summary of statistics about the datasets.
\begin{enumerate}
	\item \textbf{Clinical~\cite{malioutov2006minimum}}: Consists of a set of 227 chapters from a medical textbook. Each chapter is marked into sections indicated by the author which forms the segmentation boundaries. It contains a total of 1136 sections.
	\item \textbf{Fiction~\cite{kazantseva2011linear}}: Consists of a collection of 85 fiction books downloaded from Project Gutenberg. Segmentation boundaries are the chapter breaks in each of the books.
    \item \textbf{Wikipedia}: Consists of randomly selected set of 300 documents having an average segment size of 26. These documents widely fall under the narrative category. Each document is divided into sections as marked in the original XML dump of the website. We use these section markers to create a segmentation boundary.
\end{enumerate}

\subsection{Metrics} 
We evaluate the performance of the model with respect to two metrics: Pk~\cite{beeferman1999statistical} and WinDiff~\cite{pevzner2002critique}. Both the metrics use a sliding window of fixed size $w$ over the document and compare the hypothesized segments with the reference ones. The window size, $k$, is generally set to half the gold-standard segmentation length~\cite{beeferman1999statistical}. Pk is the probability that two segments drawn from a document are incorrectly identified as belonging to the same segment. Windiff moves a sliding window across the text and counts the number of times the hypothesized and reference segment boundaries are different within the window. Counts are scaled to obtain probability values. Both Pk and Windiff thus lie between 0 and 1 and an algorithm that assigns all boundaries correctly receives a score of 0. WinDiff is considered a better measure than Pk as the Pk metric suffers from multiple issues as described in~\cite{pevzner2002critique}. Pevzner et al.~\cite{pevzner2002critique} proposed WinDiff as an update to the Pk metric. \emph{Both the metrics are a loss measure. The lower the score, the better.}

\subsection{Model Parameters and Training}
\label{sec:model-parameters-and-training}

For the purpose of training, we randomly select a set of pages from Wikipedia. We use section splits as our segment splits for training the model and skip the section headers, considering only the section content for training. We perform a 80-20 training-development split to obtain the training and development datasets. Fig. \ref{fig:windiffIters} shows variation of WinDiff scores on the Wikipedia development dataset and other datasets as training progresses. The figure suggests that training converges well after 30 iterations, hence we fix number of iterations as 30. We trained our model on $\sim$270 documents from a sample of the Wikipedia corpus, creating $\sim$49k training sentences/samples. The average segment size of the whole Wikipedia corpus is around 9 sentences, while the test datasets have higher segment sizes (Fig. \ref{data-stats-table}). We filter the documents that have average segment size less than 20 which results in training set having average segment length of 25. We train our model in batches of size 40. We set the context size, $K$, to 10 for all the experiments mentioned in Table \ref{table:experiments:all}.

The training dataset class distribution is heavily skewed with about 92\% samples belonging to class 0. Hence, we use weighted-binary-cross-entropy as our loss function to penalize the classifier more heavily on mis-classification of a segment boundary. The loss function is defined as $loss = -\frac{1}{N} \sum_{i=1}^{N}(t \log(o) + \frac{f_1}{f_0} (1 - t) \log(1 - o)))$ where $t$ and $o$ are the target and the predicted outputs respectively. $f_0$ and $f_1$ are the frequencies of class 0 and class 1 respectively.

We use `AdaDelta' \cite{zeiler2012adadelta} as the optimizer and use dropouts of 0.2 -- 0.3 for input and recurrent gates in the recurrent layers. We also use dropouts of 0.3 after the dense fully connected layers to prevent over fitting on the training dataset. We use filters of sizes $\{2, 3, 4, 5\}$ with 200 filters for each of the sizes. The recurrent layers have 600 neurons. 

\subsection{Comparison with Other Baseline Methods}
We compare the performance of our proposed model against various competitive baselines, four basic neural models, and three BiLSTM model variants. Each of those models help us understand contributions of the various components of the proposed model. Table \ref{table:experiments:all} shows the summary of the results obtained using various models on all three benchmark datasets. In the following, we describe the baseline systems in brief.

\begin{enumerate}
	\item \textbf{U\&I} \cite{utiyama2001statistical}: It is a probabilistic framework based on maximizing the compactness of the language models induced for each segment using ideas similar to the noisy channel and minimum description length methods.
	\item \textbf{MinCut} \cite{malioutov2006minimum}: This method treats text segmentation as a graph-partitioning task aiming to optimize the normalized-cut criterion. It simultaneously optimizes the total similarity within each segment and dissimilarity across segments.
	\item \textbf{BayesSeg} \cite{eisenstein2008bayesian}: This method models the words in each topic segment as draws from a multinomial language model associated with the segment. Segmentation is obtained by maximizing the observation likelihood in such a model. 
	\item \textbf{APS} \cite{kazantseva2011linear}: Affinity Propagation for Segmentation receives a set of pairwise similarities between data points and produces segment boundaries and segment centers. Data points which best describe all other data points within the segment are considered segment centers. APS iteratively passes messages in a cyclic factor graph, until convergence. 
	\item \textbf{PLDA} \cite{Purver:2006:UTM:1220175.1220178}: PLDA is a generative model that uses Bayesian inference to simultaneously address the problems of topic segmentation and topic identification. It chains a set of LDAs by assuming a Markov structure on topic distributions.
	\item \textbf{TSM} \cite{du2013topic}: Structured Topic Model is a hierarchical Bayesian model for unsupervised topic segmentation. It uses an MCMC inference to split/merge segment(s).
\end{enumerate}

For all these baseline algorithms, we use the publicly available source codes. We also fine tune the parameters, using the scripts provided by the authors, for our experiment on the Wikipedia dataset to get the optimal set of parameters. We could not perform some experiments where source codes were not available publicly. We mark those instances with NA. We also compare the performance on ``Random'' baseline where we place the segment boundaries randomly in the text. Part A of Table \ref{table:experiments:all} shows the observed results from our experiments on these baselines.

To understand the contribution of various components in our proposed model we compare the performance of other neural models as well without using any context information. We discuss these neural models in brief below. Part B of Table \ref{table:experiments:all} presents the performance for four such neural architectures.

\begin{enumerate}
	\item \textbf{Perceptron:} We encode the sentence using mean of the word2vec representations of the corresponding words and then learn a 5-layered perceptron.
	\item \textbf{LSTM:} We represent each sentence using a sequence of words and then learn a combined dense representation for each sentence in the vector space using the word2vec embeddings for words. 
	\item \textbf{Stacked-LSTM:} This model is similar to LSTM, except that it provides more flexibility at the cost of more trainable parameters. 
	\item \textbf{CNN:} We use the CNN based sentence representations obtained by convolving multiple variable length filters with the word embeddings to obtain rich feature representations for each sentence.
\end{enumerate}

We also compare the performance of our proposed model with other BiLSTM based neural models. Each of these models obtain specific sentence representations which are then passed to a BiLSTM architecture to obtain context representations. Part C of Table \ref{table:experiments:all} presents the performance for these neural architectures besides the proposed method, CNN+Attn-BiLSTM.

\begin{enumerate}
	\item \textbf{MeanBoW-BiLSTM:} We use mean of all the word vectors to obtain a sentence representation, which along with its neighboring context, is passed to a stacked BiLSTM encoder architecture to obtain the context representations.
	\item \textbf{TFIDF MeanBoW-BiLSTM:} To obtain a sentence embedding, we compute weighted mean of the word2vec word embeddings where TF-IDF (Term Frequency-Inverse Document Frequency) scores are used as weights. TF-IDF scores capture the relative relevance of a particular word in the sentence.
    \item \textbf{CNN-BiLSTM:} We use the CNN based sentence representations obtained by convolving multiple variable length filters with the word embeddings to obtain rich feature representations for each sentence. There is no attention layer in this method.
\end{enumerate}

\begin{table}%
	\centering\scriptsize
		\begin{tabular}{|l|l|r|r|r|r|r|r|}
			\hline
			\multirow{2}{*}{} & \multirow{2}{*}{\bf Model} & \multicolumn{2}{c|}{\bf Clinical} & \multicolumn{2}{c|}{\bf Fiction} & \multicolumn{2}{c|}{\bf Wikipedia}  \\  \cline{3-8}
			& & \bf WinDiff & \bf Pk  & \bf WinDiff & \bf Pk  & \bf WinDiff & \bf Pk \\
			\hline
			\multirow{7}{*}{\parbox{1.7cm}{\textbf{Part A}:\\ Competitive Baselines}} & U\&I~\cite{utiyama2001statistical} & 0.376 & 0.370 & 0.459 & 0.459 & 0.368 & 0.368 \\
			& MinCut~\cite{malioutov2006minimum} & 0.382 & 0.368 & 0.405 & 0.371 & 0.389 & 0.364 \\
			& BayesSeg~\cite{eisenstein2008bayesian} & 0.353 & 0.339 & 0.337 & \bf 0.278 & 0.390 & 0.359 \\
			& APS~\cite{kazantseva2011linear} & 0.399 & 0.396 & 0.480 & 0.451 & 0.380 & 0.392  \\
			& PLDA~\cite{Purver:2006:UTM:1220175.1220178} & 0.373 & 0.324 & 0.430 & 0.361 & NA & NA  \\
			& TSM~\cite{du2013topic} & 0.345 & \bf 0.306 & 0.408 & 0.325 & NA & NA  \\
			& Random & 0.459 & 0.441 & 0.510 & 0.475 & 0.486 & 0.480 \\
			
			\hline
			\multirow{4}{*}{\parbox{1.7cm}{\textbf{Part B}:\\ Neural Models without Context}}
			& Perceptron & 0.338  & 0.357 & 0.336 & 0.335 & 0.421 & 0.415 \\
			& Stacked-LSTMs & 0.381 & 0.393 & 0.329 & 0.394 & 0.437 & 0.420 \\
			& LSTMs & 0.486 & 0.471 & 0.366 & 0.417 & 0.508 & 0.455 \\
			& CNN & 0.309 & 0.329 & 0.314 & 0.386 & 0.363 & 0.380 \\

			\hline
			\multirow{4}{*}{\parbox{1.7cm}{\textbf{Part C}:\\ Context based Neural Models}}  
			& MeanBoW + BiLSTM & 0.349  & 0.365 & 0.319 & 0.389 & 0.405 & 0.398 \\
			& TF-IDF MeanBoW + BiLSTM & 0.345  & 0.366 & 0.328 & 0.382 & 0.382 & 0.392 \\
			& CNN + BiLSTM & 0.334  & 0.316 & 0.331 & 0.324 & 0.378 & \bf 0.328 \\
			& \textbf{CNN + Attn-BiLSTM}& \bf 0.294  & 0.318 & \bf 0.308 & 0.378 & \bf 0.315 & 0.344 \\
			\hline
		\end{tabular}
	\caption[AAAA]{\label{table:experiments:all} \scriptsize Accuracy Comparison of the Proposed	 Approach with Competitive Baselines. \textbf{Lower values are better.} Experiments marked with \textbf{NA} could not be performed due to non-availability of publicly available source codes. Some of the cell values have been directly taken from respective papers, if they mentioned them for the same (method, dataset) pair.}
	\end{table}

\begin{figure}[!htb]
   \begin{minipage}{0.45\textwidth}
				\scriptsize
	\centering
\includegraphics[width=0.8\columnwidth]{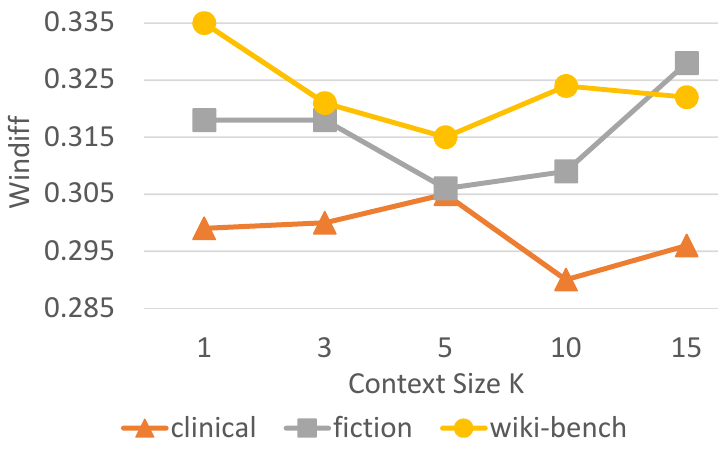}%
\caption{\scriptsize Variation of WinDiff Scores on Various Datasets with Varying Context Size K}%
\label{fig:varyKWindiff}%
\end{minipage}%
\hspace{0.04\textwidth}
   \begin{minipage}{0.45\textwidth}
				\scriptsize
	\centering
\includegraphics[width=0.8\columnwidth]{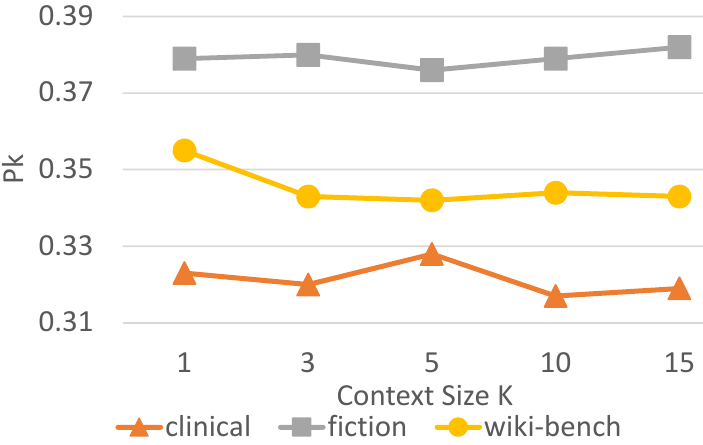}%
\caption{\scriptsize Variation of Pk Scores on Various Datasets with Varying Context Size K}%
\label{fig:varyKPk}%
    \end{minipage}%
\end{figure}

%% file: results.tex
\section{Analysis of Results}
\label{sec:results}

In this section we analyze the results of our experiments and compare their performance with the existing non-neural as well as neural baselines. We also briefly discuss about the choice of metrics in sub-section  \ref{sec:implication_of_pk}.

\subsection{Comparison with Baseline Models}

Table \ref{table:experiments:all} compares the performance of the proposed Attention-based model on various benchmark datasets with other existing models. We observe that the use of Attention-based supervised models provide a performance improvement over other methods across all the datasets on the Windiff metric, and compares well with the best method on the Pk metric.


The proposed model has additional benefits with respect to runtime performance. Once the training is finished, during the prediction phase, our model takes on an average 0.09 seconds on a batch of 40 sentences on GeForce GTX 1060 GPU which is much faster than other methods in Part A of Table \ref{table:experiments:all} which take time to the order of minutes to days during prediction phase as most of the computation takes place during that time. 

\subsection{Comparison with Neural Models}

It is important to note that without context the neural models (Part B) sometimes perform worse than the baseline models (Part A). All of the variants of LSTMs in Part C are better than the context-unaware LSTM model in Part B. In Part C, we note that the use of TFIDF-weighted-Mean BoW word2vec embeddings for representing sentences (TFIDF MeanBoW + BiLSTM) only provides a slight improvement. We conclude that for the task of text segmentation, TF-IDF features do not add additional information as compared to the word-embeddings.

Our experiments with the CNN model provides us with good results compared to other model variants overall. Results in Part B show improved performance even without using any context information with the use of CNN for obtaining sentence representations. Our experiments with CNNs along with BiLSTM do not show much improvement in the results on the Windiff metric, though they show significant improvement on the Pk metric results across all the datasets encouraging us to use BiLSTM as part of our proposed model. The use of Attention further improves the results by a significant margin across all the datasets.

We also observe improved performance with the use of Attention on the neighboring context. Use of soft Attention~\cite{xu2015show} allows the model to focus on certain regions more than others in order to generate better context representations. Using the attention layer shows an increase in Windiff performance by 4\%, 2.3\% and 6.3\% on Clinical, Fiction and Wikipedia datasets respectively over the CNN+BiLSTM model. Overall, the proposed model shows an improvement of 5.1\%, 10\% and 6.5\% on Windiff metric on Clinical, Fiction and Wikipedia datasets respectively, and improvement of 3.1\% on Pk metric on the Wikipedia dataset compared to the existing competitive baselines.

\subsection{Varying Context Size $K$}

We also experiment with the variation in context size $K$ on all the datasets and report the results in Fig.~\ref{fig:varyKWindiff} and~\ref{fig:varyKPk}. We train our model on the same training dataset for 30 epochs and report the results with variable context sizes. We observe a decreasing trend (recall low Windiff scores are good) in the WinDiff scores as Context size increases which gradually starts increasing as context size grows. All the three datasets follow this trend for the Windiff metric, while only the Fiction and Wikipedia datasets have shown similar results for Pk metric. For Clinical dataset, we lose a lot of domain specific information while converting words to vectors using the word-embeddings as word2vec is not very rich in domain specific information. This leads to poor results on the Windiff metric with low context-size. As context size grows, it gathers enough information to correctly classify the segment boundaries. 



\subsection{Implication of Windiff and Pk Metric Results}
\label{sec:implication_of_pk}

Since the models are trained on the Wikipedia Dataset, it is focused towards presenting an evenly spread out segmentation of the paragraph, as learned from the Wikipedia documents, which often have a uniform section distribution. The Clinical and Fiction datasets both have quite a significant number of segments with less number of sentences (as evident from the very high standard deviation reported in Table \ref{data-stats-table}) resulting in less number of predicted segment boundaries.
Assuming a window of size $\lambda$, each miss of segmentation boundary will produce a false negative. Each such false negative will receive a total of $\lambda$ penalties. Since the model often results in less number of segmentations, it often receives higher number of penalties than expected, resulting in higher Pk scores. Hence, as also seen in Table \ref{table:experiments:all}, baseline neural models consistently perform poorly on the Pk metric. But the Pk scores for neural models on Wikipedia dataset are comparable to the state-of-the-art methods.

\underline{Choice of WinDiff over Pk}: Pk metric suffers from multiple issues due to which it does not provide a good measure of the hypothesized segmentations. These issues are covered in detail by Pevzner et al.~\cite{pevzner2002critique}, motivating us to use the WinDiff metric as our primary measure for comparing performance with the existing models and baselines. We still report the results in both the metrics as they might provide beneficial information to the readers.

%% file: conclusion.tex
\section{Conclusions}
\label{sec:conclusion}

In this paper, we studied the problem of text segmentation from a neural perspective. We presented a model which first learns rich features for every sentence using Convolutional Neural Networks followed by sequential learning using temporal data. Finally, we also learn focus on various sentences in the context using the attention layer. We performed extensive experiments to compare against well-established non-neural baselines, as well as against recent neural models. Experimenting with three different datasets, we empirically proved that our proposed model provides lowest Windiff loss with very little supervision, and with low execution times.


In the future, we plan to test this model on non-English datasets, especially morphologically rich languages where huge datasets are not available for training the model and most of the contextual information is captured at the word level rather than at the sentence level.